\documentclass{ecai}
\usepackage{times}
\usepackage{latexsym}

\usepackage{microtype}      \usepackage{nicefrac}       \usepackage[utf8]{inputenc}
\usepackage[T1]{fontenc}

\usepackage{booktabs,multirow}
\usepackage[table]{xcolor}
\newcolumntype{C}{c<{\kern\tabcolsep}@{}}
\newcolumntype{R}{r<{\kern\tabcolsep}@{}}
\newcolumntype{X}{r<{\scalefont{.9} $\pm$ }@{}}
\newcolumntype{Y}{>{\scalefont{.9}}r<{\kern\tabcolsep}@{}}

\usepackage{rotating}

\usepackage{tikz}
\usetikzlibrary{shapes.geometric}
\usetikzlibrary{shapes}
\usetikzlibrary{shapes.multipart}

\usepackage[pdftex]{pstricks}
\usepackage{verbatim,moreverb}
\usepackage{url}
\usepackage{booktabs}       \usepackage{enumitem}

\usepackage{xspace}
\usepackage{graphicx}
\usepackage{subfig}
\graphicspath{{./}{./Images/}{./figures/}}

\usepackage{multicol} \usepackage{siunitx}

\usepackage[numbers]{natbib}
\renewcommand{\citep}[1]{\cite{#1}}
\renewcommand{\citet}[1]{\citeauthor{#1} \cite{#1}}

\usepackage{pifont}

\usepackage[normalem]{ulem}
\usepackage{scalefnt}
\usepackage{amsmath}
\usepackage{amssymb}
\usepackage{mathtools}
\usepackage{empheq}
\usepackage{cleveref}
\expandafter\def\csname ver@etextools.sty\endcsname{3000/12/31}
\expandafter\def\csname ver@etex.sty\endcsname{3000/12/31}

\usepackage{autonum}
\allowdisplaybreaks
\usepackage{centernot}

\DeclareMathOperator*{\argmax}{arg\,max}
\def\ie{{\em i.e.}\xspace}
\def\eg{{\em e.g.}\xspace}
\def\cf{{\em cf.}\xspace}

\def\cS{{\cal S}}
\def\cA{{\cal A}}

\def\cB{{\cal B}}
\def\cZ{{\cal Z}}
\def\cH{{\cal H}}

\def\bag{\beta}
\def\depth{\delta}

 \def\Pazba[#1]{b^{a,z}_{#1}}
\def\Pazbb[#1]{\frac{P_{a,z}b_{#1}}{\norm{P_{a,z}b_{#1}}_1}}

\usepackage[ruled,vlined]{algorithm2e}
\SetKwProg{Fct}{Fct}{}{}

\newtheorem{proof}{Proof}

\newcommand{\eqdef}     {\doteq}

\newcommand{\qed}{\hfill  $\square$} 

\DeclarePairedDelimiter{\norm}{\lVert}{\rVert}

\ifdefined\nocomments
  \newcommand{\Vincent}[1]{}
  \newcommand{\Olivier}[1]{}
  \renewcommand{\sout}[1]{}
  \renewcommand{\xout}[1]{}

\else
  \newcommand{\Vincent}[1]{\textcolor{purple}{[vt] #1}}
  \newcommand{\Olivier}[1]{\textcolor{blue}{[ob] #1}}   
\fi

\newcommand{\tcr}[1]{\textcolor{red}{#1}}
\definecolor{darkgreen}{rgb}{0, 0.35, 0}

\begin{document}

\title{Monte Carlo Information-Oriented Planning\\
(Revised version)}

\author{Vincent Thomas
\institute{Universit\'e de Lorraine, INRIA, CNRS, LORIA, France,
vincent.thomas@loria.fr}
\and G\'er\'emy Hutin
\institute{\'Ecole Normale Sup\'erieure de Lyon, France,
geremy.hutin@ens-lyon.fr}
\and Olivier Buffet
\institute{Universit\'e de Lorraine, INRIA, CNRS, LORIA,
France,
olivier.buffet@loria.fr}
}

\maketitle

\def\myecai{ECAI}
\newcommand{\betasize}{$\vert \beta \vert$\xspace}

\newcommand{\newFinal}[1]{\textcolor{blue}{#1}}

\ifdefined\myecai
\begin{abstract}
  In this article, we discuss how to solve information-gathering problems
  expressed as $\rho$-POMDPs, an extension of Partially Observable
  Markov Decision Processes (POMDPs) whose reward $\rho$
  depends on the belief state.
  Point-based approaches used for solving POMDPs have been extended to
  solving $\rho$-POMDPs as belief MDPs when its reward $\rho$ is
  convex in $\cB$ or when it is Lipschitz-continuous.
  In the present paper, we build on the POMCP algorithm to
  propose a Monte Carlo Tree Search for $\rho$-POMDPs, aiming for an
  efficient on-line planner which can be used for any $\rho$ function.
  Adaptations are required due to the belief-dependent rewards to (i)
  propagate more than one state at a time, and (ii) prevent biases in
  value estimates.
  An asymptotic convergence proof to $\epsilon$-optimal values is given when $\rho$ is continuous.
  Experiments are conducted to analyze the algorithms at hand and show
  that they outperform myopic approaches.
\end{abstract}
\fi

\paragraph{Preliminary note.} 
This article is a revised version of the ECAI 2020 paper.
This version better highlights the differences between our contribution, renamed $\rho$-POMCP($\beta$), and the original $\rho$-POMCP algorithm \cite{Bargiacchi16}.
We thus recommend reading and citing this revised version rather than the original ECAI 2020 paper.

\section{INTRODUCTION}

Many state-of-the-art algorithms for solving Partially Observable
Markov Decision Processes (POMDPs) rely on turning the problem into a
``fully observable'' problem---namely a belief MDP---and exploiting
the piece-wise linearity and convexity of the optimal value function
in this new problem's state space (here the belief space $\cB$) by
maintaining generalizing function approximators.
This approach has been extended to solving
$\rho$-POMDPs as belief MDPs---\ie, problems whose performance
criterion depends on the belief (\eg, active information
gathering)---when the reward $\rho$ itself is convex in $\cB$
\citep{AraBufThoCha-nips10} or when $\rho$ is Lipschitz-continuous \citep{FehBufThoDib-nips18}.

In this paper, we propose two new algorithms for solving $\rho$-POMDPs
which do not rely on properties of $\rho$ such as its convexity or its Lipschitz-continuity,
but are based on Monte Carlo sampling, are inspired by POMCP \citep{SilVen-nips10}
and, while developed independently, extend the general direction proposed by \citet{Bargiacchi16}.
These algorithms are :
(1) \textit{$\rho$-beliefUCT}, which applies UCT to the belief MDP; and
(2) \textit{$\rho$-POMCP$(\beta)$}, which uses particle filters, \ie, samples particles during trajectories, to estimate the visited belief states
and their associated rewards.
We prove $\rho$-POMCP$(\beta)$'s asymptotic convergence and empirically assess
these algorithms on various information-gathering problems.

The paper is organized as follows.
Section~\ref{sec:relatedwork} discusses related work on information-oriented control.
Sec.~\ref{sec:background} presents background
(i) first on POMDPs,
(ii) then on the Partially Observable Monte Carlo Planning (POMCP) algorithm, a Monte Carlo Tree Search approach for solving POMDPs, and
(iii) finally on $\rho$-POMDPs.
Sec.~\ref{sec:MCTSrho} describes our contribution:
$\rho$-beliefUCT and $\rho$-POMCP$(\beta)$, for solving $\rho$-POMDPs with MCTS techniques, and provides a proof of $\rho$-POMCP$(\beta)$'s convergence.
Sec.~\ref{sec:XPs} presents the conducted experiments and analyzes the results before concluding and giving some perspectives.

\section{RELATED WORK}
\label{sec:relatedwork}

Early research on information-oriented control (IOC) involved problems
formalized either
(i) as POMDPs (as \citet{EgoKocUud-aaai16} did recently, since an
observation-dependent reward can be trivially recast as a
state-dependent reward), or
(ii) with belief-dependent rewards (and mostly ad-hoc solution
techniques as \citep{FoxBurThr-ras98,MihLefBruSch-nato06}).

\citet{AraBufThoCha-nips10} introduced $\rho$-POMDPs to easily
formalize most IOC problems.
They showed that a $\rho$-POMDP with convex belief-dependent reward
$\rho$ can be solved with modified point-based POMDP solvers
exploiting the piece-wise linearity and convexity (PWLC property), with error bounds that depend on the
quality of the PWLC-approximation of $\rho$.
More recently, \citet{FehBufThoDib-nips18} applied the same approach
as \citeauthor{AraBufThoCha-nips10}, but for Lipschitz-continuous
(LC)---rather than convex---belief-dependent rewards,
demonstrating that, for finite horizons, the optimal value function is
also LC.
Then, deriving uniformly improvable lower- (and upper\mbox{-)}bounds
led to two algorithms based on HSVI \citep{Smith-phd07}.

\citeauthor{SpaVeiLim-jaamas15}'s POMDP-IR framework
allows describing IOC problems with linear
rewards which are provably equivalent to ``PWLC'' $\rho$-POMDPs (\ie,
when $\rho$ is PWLC) \citep{SatWhiSpa-tr15}, and also leads to
modified POMDP solvers.
For its part, the general $\rho$-POMDP framework allows formalizing
more problems---\eg, directly specifying an entropy-based criterion.

We here consider $\rho$-POMDPs, but not relying on generalizing (PWLC
or LC) value function approximators as in previous work,
so that any belief-dependent reward can be used.
For instance, this allows addressing problems where the objective is to minimize the quantity of information of an adversary with known behaviour
or where the objective is to gather information on specific state variables while maintaining a low quantity of information on confidential ones (like in medical or domotic fields).
None of these problems can be solved by previous approaches on $\rho$-POMDP nor be modelled by POMDP-IR, which requires a PWLC reward function in $\cB$.
To circumvent this difficulty, we build on Monte Carlo Tree Search (MCTS) approaches,
in particular
\citeauthor{SilVen-nips10}'s Partially Observable Monte Carlo Planning
(POMCP) algorithm \citeyear{SilVen-nips10}.
Moreover, MCTS and POMCP present several other benefits:
(i) they require a simulator rather than a complete model;
(ii) unlike heuristic search, they do not require
optimistic or pessimistic initializations of the value function;
and (iii) they have been proven to be very efficient for solving
problems with huge state, action and observation spaces.

\citeauthor{Bargiacchi16}'s work \cite{Bargiacchi16}, while developed
independently from the present work, already proposed
(i) the same key idea of applying POMCP on $\rho$-POMDPs, thus deriving the same
global algorithmic scheme as our contribution (\cf
Sec.~\ref{sec:rho-POMDP}), hence the name: $\rho$-POMCP, and
(ii) two variants also presented in this article but which  we discuss in more details.
Note that our main algorithm is one of Bargiacchi's variants, and reciprocally.
However, the present paper differs from \citeauthor{Bargiacchi16}'s work and extends it in that it:
(i) explicitly adds a particle filter to improve the belief estimate, 
(ii) evaluates the proposed algorithm and the two variants (while Bargiacchi's evaluation concentrates on his main algorithm), 
(iii) conducts experiments on larger and more diverse problems,
(iv) provides theoretical convergence guarantees for generic $\rho$ functions,
and (v) discusses $\rho$'s continuity assumption (which is required for
proving asymptotic convergence).
We believe that these results demonstrate the value of MCTS approaches
for information-oriented control, and that our work provides a more
in-depth (theoretical and empirical) look into them.

\citet{Lauri16} also propose a POMCP algorithm for
information-oriented \textit{open-loop} planning, \ie,
finding an action sequence for a $\rho$-POMDP while ignoring
observations for action selection during planning.
While addressing {continuous POMDPs} (\ie, with continuous or numerous states/actions/observations but with a reward defined over state and action), \citeauthor{SunKoc-corr18} \cite{SunKoc-corr18,SunKoc-icaps18} also combine POMCP with particle filters and point out that their approach can be applied to $\rho$-POMDPs, opening new directions for continuous $\rho$-POMDPs.

\section{BACKGROUND}
\label{sec:background}

\subsection{POMDPs}

A POMDP \citep{Astrom-jmaa65} is defined by a tuple
$\langle \cS, \cA, \cZ, P, r, \gamma, b_0 \rangle$, where
$\cS$, $\cA$ and $\cZ$ are finite sets of states, actions and
observations;
$P_{a,z}(s,s')$ gives the probability of transiting to state $s'$ and
observing observation $z$ when applying action $a$ in state $s$
($P_{a,z}$ is an $\cS\times\cS$ matrix);
$r(s,a) \in \mathbb{R}$ is the reward associated to performing action
$a$ in state $s$;
$\gamma \in [0,1)$ is a discount factor; and
$b_0$ is the initial belief state---\ie, the initial probability
distribution over possible states.
The objective is then to find a policy $\pi$ that prescribes actions
depending on past actions and observations so as to maximize the
expected discounted sum of rewards (here with an infinite temporal
horizon).

To that end, a POMDP is often turned into a belief MDP
$\langle \cB,\cA,P,r,\gamma,b_0 \rangle$, where
$\cB$ is the belief space,
$\cA$ is the same action set, and
$P_a(b,b')=P(b'|b,a)$ and $r(b,a)=\sum_{s \in \cS}  b(s)r(s,a)$ are the induced
transition and reward functions.
This allows considering policies $\pi \colon \cB \to \cA$, and their value functions
$V^\pi(b)\eqdef E [\sum_{t=0}^\infty \gamma^t r(b_t,\pi(b_t)) | b_0=b ]$.
Optimal policies maximize $V^\pi$ in all belief states reachable from
$b_0$.
Their value function $V^*$ is the fixed point of Bellman's {\em
  optimality} operator ($\cH$) \citep{Bellman-jmm57}
$ \cH V: b \mapsto \max_a [ r(b,a) + \gamma \sum_z \|P_{a,z} b\|_1
V(\Pazba[])]$, and
acting greedily with respect to $V^*$ provides such a policy.

\subsection{MCTS for POMDPs}

\paragraph{MCTS and UCT}
MCTS approaches \citep{Coulom-cg06,KocSze-ecml06,BroEtAl-tciaig12} are
online, sampling-based algorithms, here described in the MDP framework
(while they also serve in settings like sequential games).
In MCTS, the tree representing possible futures from a starting state
is progressively built by sampling trajectories in a non-uniform way.
Each iteration consists in 4 steps:
{\bf (i) selection}: a trajectory is sampled in the tree according to an exploration strategy until a node not belonging to the tree is reached;
{\bf (ii) expansion}: this new node is added to the tree;
{\bf (iii) simulation}: this new node's value is estimated by sampling a trajectory from this node according to a rollout policy;
{\bf (iv) backpropagation}: this estimate and the rewards received during the selection step are back-propagated to the visited nodes to update their statistics (value and number of visits).
Upper Confidence Bound applied to trees (UCT) is an instance of MCTS
where, when in a state node, the next action is selected using the
Upper Confidence Bounds (UCB1) strategy, \ie, picking an action so as
to maximize its estimated value increased by an exploration bonus
$c_{t,s} = C_p \sqrt{{\log N(s)}\over{N(s,a)}}$,
with
$C_p>0$ an exploration constant, $N(s)$ the number of past visits of node $s$, and
$N(s,a)$ the number of selections of action $a$ when in node $s$.

\begin{figure}
  \center
    \resizebox{\columnwidth}{!}{\tikzset{
itria/.style={
  draw,dashed,shape border uses incircle,
  isosceles triangle,shape border rotate=90,yshift=-0.1cm},
rtria/.style={
  draw,dashed,shape border uses incircle,
  isosceles triangle,isosceles triangle apex angle=90,
  shape border rotate=-45,yshift=0.2cm,xshift=0.5cm},
ritria/.style={
  draw,dashed,shape border uses incircle,
  isosceles triangle,isosceles triangle apex angle=110,
  shape border rotate=-55,yshift=0.1cm},
letria/.style={
  draw,dashed,shape border uses incircle,
  isosceles triangle,isosceles triangle apex angle=110,
  shape border rotate=235,yshift=0.1cm}
}

\begin{tikzpicture}

  \tikzstyle{every text node part}=[align=center]

  \node(test)[rectangle, draw]{Root Belief $h=()$\\ $b_0$} [sibling distance=6cm]
    child {
      node [draw, ellipse]{$ha=(a^1)$ \\ $V(ha),N(ha)$} [sibling distance=3cm, level distance=2cm]
      child {
        node [rectangle, draw]{$h=(a^1,z^1)$ \\ $B(h),V(h),N(h)$} [sibling distance=1.5cm, level distance=1.5cm]
        child {
          node [circle, draw]{} [level distance=1.3cm]
          child {
            node[itria]{$\ldots$}
          }
          edge from parent node{$a^1$}
          }
        child {
          node [circle, draw]{} [level distance=1.3cm]
          child {
            node[itria]{$\ldots$}
          }
          edge from parent node{$a^2$}
          }
        edge from parent node{$z^1$}
      }
      child {
        node [rectangle, draw]{$h=(a^1,z^2)$ \\ $B(h),V(h),N(h)$} [sibling distance=1.5cm, level distance=1.5cm]
        child {
          node [circle, draw]{} [level distance=1.3cm]
          child {
            node[itria]{$\ldots$}
          }
          edge from parent node{$a^1$}
          }
        child {
          node [circle, draw]{} [level distance=1.3cm]
          child {
            node[itria]{$\ldots$}
          }
          edge from parent node{$a^2$}
        }
        edge from parent node{$z^2$}
      }
      edge from parent node{$a^1$}
    }
    child {
      node [draw, ellipse]{$ha=(a^2)$ \\ $V(ha),N(ha)$} [sibling distance=3cm, level distance=2cm]
      child{
        node [rectangle, draw]{$h=(a^2,z^1)$ \\ $B(h),V(h),N(h)$} [sibling distance=1.5cm, level distance=1.5cm]
        child {
          node [circle, draw]{} [level distance=1.3cm]
          child {
            node[itria]{$\ldots$}
          }
          edge from parent node{$a^1$}
          }
        child {
          node [circle, draw]{} [level distance=1.3cm]
          child {
            node[itria]{$\ldots$}
          }
          edge from parent node{$a^2$}
          }
        edge from parent node{$z^1$}
      }
      child{
        node [rectangle, draw]{$h=(a^2,z^2)$ \\ $B(h),V(h),N(h)$} [sibling distance=1.5cm, level distance=1.5cm]
        child {
          node [circle, draw]{} [level distance=1.3cm]
          child {
            node[itria]{$\ldots$}
          }
          edge from parent node{$a^1$}
          }
        child {
          node [circle, draw]{} [level distance=1.3cm]
          child {
            node[itria]{$\ldots$}
          }
          edge from parent node{$a^2$}
          }
        edge from parent node{$z^1$}
        edge from parent node{$z^2$}
      }
      edge from parent node{$a^2$}
    };
\end{tikzpicture}
 }
  \captionsetup{justification=centering}
  \caption{Example of a belief tree with 2 actions: $a^1$ and $a^2$, and 2 observations: $z^1$ and $z^2$, with various quantities maintained by POMCP.
  }
  \label{fig:belieftree}
\end{figure}
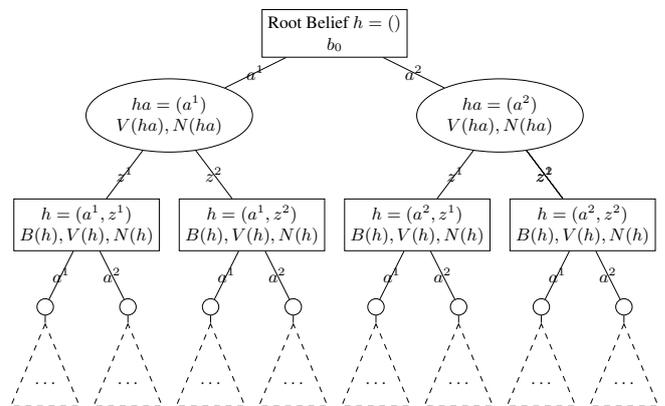

\paragraph{POMCP}
\citet{SilVen-nips10}
proposed the Partially Observable Monte Carlo Planning (POMCP) algorithm to apply MCTS for solving POMDPs.
A POMDP is addressed through the corresponding belief MDP, a belief tree being made of alternating action and belief nodes as presented in Figure~\ref{fig:belieftree}.
The path to a belief node at depth $t$ follows the action-observation history  $h_t=(a_0,z_0,a_1,z_1,\dots,z_t)$  leading from the root belief to this belief $b(h_t)$.

Applying directly UCT on the belief-MDP would imply sampling trajectories $(b_0,a_0,b_1,a_1,\dots,b_t)$ in belief space, thus requiring the complete POMDP model and a high computational cost to derive exact belief states.
To prevent this cost, POMCP samples trajectories in state space, which only requires a simulator $\mathcal{G}$ as a generative model of the POMDP.
During the selection phase, the first state is sampled from the root belief estimate, then one alternates between
  (i) picking an action according to UCB1 (applied to
  estimated action values in the current belief node), and
  (ii) sampling a next state $s'$, observation $z$ and reward $r$ using $\mathcal{G}(s,a)$.
By accumulating values in the belief nodes, averaging over all simulated trajectories gives an estimate of the value $V(h)$ of the belief node $h$.

Moreover, states are collected in each visited belief node in order to estimate the next root belief when an action is actually performed.
By preventing the computation of exact beliefs $b(h)$ in each belief node, POMCP allows addressing large problems while preserving UCT's asymptotic convergence.

\citeauthor{SilVen-nips10} also proposed the PO-UCT algorithm as a first step towards POMCP.
It differs from POMCP in that it does not collect states, but computes the belief state of the new root with an exact Bayes update.

\subsection{$\rho$-POMDPs}
\label{subsec:rhoPOMDP}

$\rho$-POMDPs \citep{AraBufThoCha-nips10} differ from POMDPs in that
their reward function $\rho$ is belief-dependent, thus allowing to define not
only control-oriented criteria, but also information-oriented ones,
thus generalizing POMDPs.
The immediate reward $\rho$ is naturally defined as $\rho (b,a,b')$, the immediate reward associated to transiting from belief $b$ to belief $b'$ after having performed action $a$.\footnote{  Without loss of generality, the article uses $\rho (b,a)$ (even if the addressed problems rely on $\rho (b,a,b')$).
}

As presented in related work, \citet{AraBufThoCha-nips10} and \citet{FehBufThoDib-nips18}
have exploited PWLC and Lipschitz-continuous reward functions $\rho$ to solve general $\rho$-POMDPs.
However, while many problems can be modeled with convex or Lipschitz-continuous $\rho$, this leaves us
with a number of problems that cannot be solved with similar approximations or with the POMDP-IR approach
(for instance, when we seek to minimize information or to find a compromise between gathering information and preserving privacy as presented in Sec. \ref{sec:relatedwork}).

Here, we propose to use the MCTS approach to address the general
$\rho$-POMDP case with no constraints on the $\rho$ function.
As an example, let us consider an agent monitoring a museum and whose aim is to locate a visitor with a specified certainty.
If $X$ denotes the random variable for the visitor's location and $b_X$ the corresponding belief,
then the reward function
$\rho_X(b,a) \eqdef \mathbf{1}_{\norm{b_X}_\infty > \alpha}$\footnote{with $\mathbf{1}$ denoting the indicator function.}
rewards beliefs whose maximum probability is greater than $\alpha\in [0,1]$.
This is a threshold function, thus neither convex nor Lipschitz-continuous, due to its
discontinuity when $\norm{b_X}_\infty = \alpha$.

\section{MCTS ALGORITHMS FOR $\rho$-POMDPs}
\label{sec:MCTSrho}

POMCP and its variant PO-UCT cannot be applied directly to solve $\rho$-POMDPs.
PO-UCT does not compute exact beliefs during its selection and backpropagation steps
(only at the tree root), and thus cannot compute belief-dependent rewards along trajectories.
POMCP generates trajectories following a single sequence of states
and samples associated rewards.
It collects visited states in belief nodes, which are not sufficient to correctly estimate the belief and thus belief-dependent rewards.

\subsection{$\rho$-beliefUCT}

The first proposed algorithm, \textit{$\rho$-beliefUCT},
consists in directly applying UCT to the belief MDP built from the $\rho$-POMDP.
This requires accessing the complete $\rho$-POMDP model and computing exact beliefs for each visited belief node
by performing Bayes updates:
$b_{haz}(s') \propto \sum_{s \in S} P_{a,z}(s,s').b_h(s)$.

During the simulation step, a belief-based rollout policy (in contrast with a random one) needs to compute the belief at each action step not only to make action choices, but also to compute immediate rewards $\rho(b,a)$ and estimate the value of the added node.
This induces an important computational cost.
However, $\rho$-beliefUCT can take advantage of several elements.
First, since each history corresponds to a unique belief state, each belief state needs to be computed only once, whenever a new belief node is added.
Secondly, while, in a POMCP approach, $r(b,a)$ is estimated using averages
of r(s,a)
without bias
because $r(b,a)$ is linear in $b$, in our context, $\rho(b,a)$ is a deterministic function of the belief:
its exact value can be directly stored in the action node.
Lastly, during the selection step, an observation can be sampled without computing observation probability $P(z|b(h))$ and without bias by sampling a state $s$ from current belief $b(h)$, then a state-observation pair $(s',z)$ from $\mathcal{G}$.

$\rho$-beliefUCT is an interesting reference algorithm.
As a direct adaptation of UCT, it inherits its convergence
properties only assuming that $\rho$ is bounded \citep{GriValMun-nips16}.
However, to avoid
(i) the cost associated to exact belief computations, and
(ii) the need for the complete model of the POMDP,
we also propose another algorithm, $\rho$-POMCP$(\beta)$, which does not compute exact beliefs but estimates them.

\subsection{$\rho$-POMCP$(\beta)$}
\label{sec:rho-POMDP}

As a first approximation, the $\rho$-POMCP$(\beta)$ algorithm is similar to POMCP.
During the selection step, trajectories are generated by sampling states and observations using generative model $\cal G$.
When visited, each belief node $h$ collects the state that has led to this node
in order to build an estimate of the true belief state $b(h)$.

However, applying directly POMCP by only adding the current state of the trajectory to the belief node, as proposed in the original $\rho$-POMCP algorithm \cite{Bargiacchi16}, may lead to poorly estimated immediate rewards during the first steps of the algorithm, thus causing the MCTS algorithm
(i) to unefficiently spend time focusing on branches with over-estimated rewards and
(ii) to slow down the exploration of branches with under-estimated rewards.

\begin{figure}  \includegraphics[width=\columnwidth]{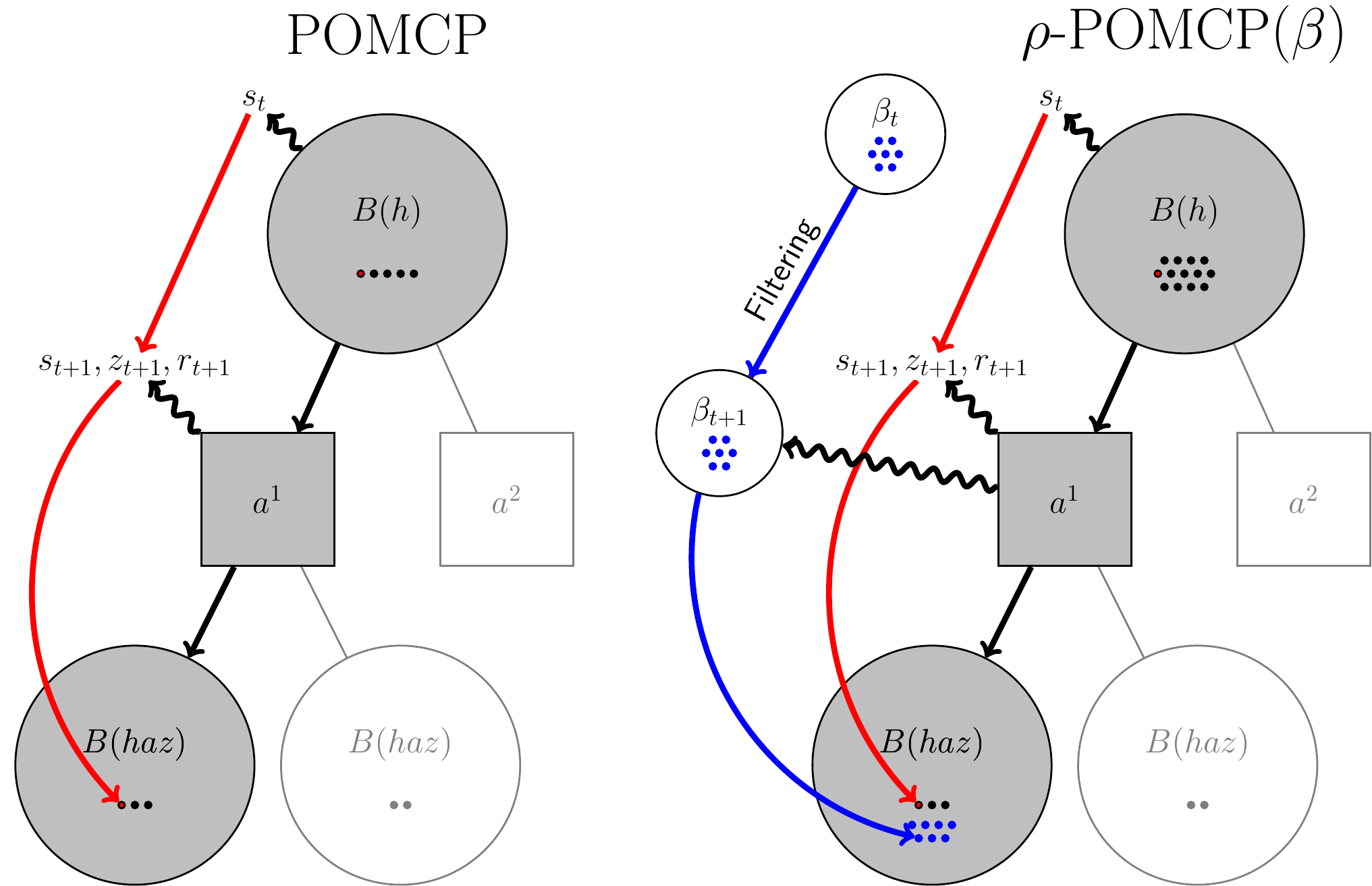}
  \captionsetup{justification=centering}
  \caption{Difference between a POMCP (or $\rho$-POMCP) descent and a $\rho$-POMCP$(\beta)$ descent: during a $\rho$-POMCP$(\beta)$ descent, at each simulation step, a bag of particles $\beta_{t+1}$ is generated from the bag $\beta_t$, the selected  action and the received observation. The particles of this bag $\beta_{t+1}$ are added to the node $B(h_{t+1})$ to produce a better estimation of $b(h_{t+1})$ than with only the sampled trajectory state.
  }
  \label{fig:rhoPOMCP}
\end{figure}

To add more states at each visit of a belief node we propose in
  $\rho$-POMCP$(\beta)$ to explicitely use a particle
filter, as illustrated in Figure~\ref{fig:rhoPOMCP}.
Thus, at each transition, when action $a_t$ is selected from trajectory state $s_t$ and observation $z_{t+1}$ is sampled,
the algorithm generates a set $\beta_{t+1}$ of $\vert \beta \vert$ states, called a \textit{small bag of particles}, using $\mathcal{G}$ to make $\vert \beta \vert$ samplings consistent with observation $z_{t+1}$.
Then, it adds this small bag $\beta_{t+1}$ to $B(haz)$,\footnote{with $a=a_t$ and $z=z_{t+1}$} the \textit{cumulative bag} for history $haz$, stored in the corresponding belief node: $B(haz) \gets B(haz) \cup \beta_{t+1}$.
It must be noted that $\rho$-POMCP$(0)$ consists in adding, at each visited belief node, only the state of the trajectory that led to this node, and thus corresponds to the initial $\rho$-POMCP algorithm proposed in \cite{Bargiacchi16}.

\paragraph{Rejection sampling}
With a generative model $\mathcal{G}$, only rejection approaches (also referred as logic sampling \cite{Henrion88}) can be considered to produce consistent samples.
In this case, a pair $(\tilde s',\tilde z)$ is sampled from ${\cal G}(s, a)$,
and $\tilde s'$ is kept in $\beta$ if and only if $\tilde z$
corresponds to the observation $z$ of the sampled trajectory.
This process repeats until enough consistent samples have been obtained,
which can be computationally expensive when $P(z|h,a)$ is small.

\paragraph{Importance sampling}
To avoid the computational cost of rejection sampling, we instead use importance sampling \citep{Doucet2000}, which assumes that the observation function is available and leads to weighting each particle.
After performing action $a_t$ and receiving observation $z_{t+1}$, a new small bag $\beta_{t+1}$ is generated from $\beta_t$ by using the generative model and following these steps:
(1) sample a state $\tilde s$ from $\beta_t$;
(2) sample a state $\tilde s'$ by using the generative model, $\tilde s' \sim {\cal G}(\tilde s, a_t)$;
(3) store this particle $\tilde s'$ in $\beta_{t+1}$ with an associated weight of $P(z_{t+1}|\tilde s,a_t,\tilde s')$ corresponding to the probability of having generated observation $z_{t+1}$.
The weights in two small bags associated to the same node can be compared, and one can thus accumulate such bags in the corresponding belief node.
To save memory, when a small bag $\beta$ is added to a
\textit{cumulative bag} $B(h)$, particles corresponding to the same
state are merged, their weights being added up.
Note that, to prevent particle depletion, the state $s_{t+1}$ in trajectory $h_t=(s_0,a_0,\dots,a_{t},s_{t+1})$ is also included in small bag $\beta_{t+1}$ with its corresponding weight, so that the observation generated during the sampled trajectory can always be explained by at least one particle in $\beta_{t+1}$.
While this induces a bias in this trajectory's small bags, since
trajectory states are obtained by sampling from the root belief,
the distribution of the states in the \textit{cumulative
  bags} $B(h)$ is unbiased.

In the simulation step, belief estimates are required at least to
estimate instant rewards, and possibly to make decisions.
The sequence of small bags $\beta$ thus needs to be perpetuated.

Finally, during the back-propagation step, the \textit{cumulative bag} $B(h)$ of belief
node $h$ is used to estimate true belief state $b(h)$---by normalizing the weights---and compute $\rho(B(h))$.
Weights are stored un-normalized in $B(h)$ so that new weighted
particles can be added without introducing a bias.

The above algorithm description leads to Algorithm~\ref{alg:rhoPOMCP}.
In this algorithm, the notation $T$ corresponds to the belief tree and $I$ to the initial belief.
$s, \bag \overset{1+n}{\sim} I$ indicates that $1+n$ states are
sampled from $I$, $1$ being stored in $s$ and the $n$ others in
$\bag$.
The function \textit{PF} corresponds here to the importance sampling particle filtering process previously described and $w_{s'}=P(z|s,a,s')$.

\SetKwFunction{estimateFct}{\textsc{Estimate}}

\DecMargin{0.5em}
\begin{algorithm}
  \caption{$\rho$-POMCP$(\beta)$     \newline {\scriptsize Text in \tcr{red} highlights differences
      with POMCP.}     \newline
    \scriptsize $\rho$-POMCP$(0)$=$\rho$-POMCP \cite{Bargiacchi16}: \ie, w/o
      particle filter, setting $\bag'=\{(s')\}$. \newline
    Note that implementation details may differ.   }  \label{alg:rhoPOMCP}
  \DontPrintSemicolon

  \SetKwFunction{searchFct}{\textsc{Search}}
  \SetKwFunction{rolloutFct}{\textsc{Rollout}}
  \SetKwFunction{simulateFct}{\textsc{Simulate}}

  \setlength{\columnsep}{13pt}   \begin{multicols}{2}

    \Fct{\searchFct$(h)$}{

      \Repeat{{\sc TimeOut}$()$}{
        \If{$h = empty$}{
          \tcr{$s, \bag \overset{1+n}{\sim} I$}
        }\Else{
          \tcr{$s, \bag \overset{1+n}{\sim} B(h)$}
        }
        \simulateFct$(s, \tcr{\bag,}\ h, 0)$      }
      \Return{$\argmax_b V(hb)$}
    }

    \vfill

    \Fct{\rolloutFct$(s, \tcr{\bag,}\ h, \depth)$}{
      \lIf{$\gamma^{\depth} < \epsilon$}{
        \Return{$0$}
      }
      $a \sim \pi_{rollout} (h, \cdot)$\;
      $(s', z) \sim {\cal G}(s, a)$\;
      \tcr{$\bag' \gets $\textsc{PF}$(\bag, a, z) \newline
        ~ \qquad \cup \{(s',w_{s'})\}$}\;
      \Return{$\tcr{\rho(\bag,a)} \newline
        ~ \qquad + \gamma.$\rolloutFct$ \newline
        ~ \hfill (s', \tcr{\bag',}\ haz, \depth+1)$}
    }

    \Fct{\simulateFct$(s, \tcr{\bag,}\ h, \depth)$}{
      \lIf{$\gamma^{\depth} < \epsilon$}{
        \Return{$0$}
      }
      \If{$h \not\in T$}{
        \lForAll{$a\in \cA$}{
          $T (ha) \gets \newline
          ~ \qquad (N_{init} (ha), \newline
          ~ \qquad V_{init} (ha), \emptyset)$
        }
        \Return{\rolloutFct$(s, \tcr{\bag,}\ h, \depth)$}
      }
      $a \gets \argmax_b V(hb) + \newline
      ~ \hfill c \sqrt{\frac{\log N(h)}{N(hb)}}$\;
      $(s', z) \sim {\cal G}(s, a)$\;
      \tcr{$\bag' \gets $\textsc{PF}$(\bag, a, z) \newline
        ~ \qquad \cup\{(s',w_{s'})\}$}\;
      $B(h) \gets B(h) \cup \tcr{\bag}$\;
      $R \gets \tcr{\rho(B(h),a)} \newline
      ~ \qquad + \gamma . $\simulateFct$ \newline
      ~ \hfill (s', \bag' , haz, \depth + 1)$\;
      $N(h) \gets N(h) + 1$\;
      $N(ha) \gets N(ha) + 1$\;
      $V(ha) \gets \newline
      ~ \qquad V(ha) + \frac{R-V(ha)}{N(ha)}$\;
      \Return{$R$}
    }

  \end{multicols}

  \medskip

\end{algorithm}
\IncMargin{0.5em}

Note that, as in POMCP or in standard particle filters, when the system actually evolves,
the actual observation may not be explained by any state contained in the new root's cumulative bag.
We leave this case for discussion, but it is still possible to re-estimate the current belief
by simulating the process from the initial belief.

\paragraph{Asymptotic convergence}

The following theorem states the asymptotic convergence of $\rho$-POMCP$(\beta)$, a notable difference with $\rho$-beliefUCT being the need for $\rho$ to be continuous in $b$ and bounded by $\rho_{max}$.
This convergence proof holds in particular for $\rho$-POMCP$(0)$, thus answering an open question by demonstrating that the original $\rho$-POMCP \cite{Bargiacchi16} asymptotically converges to an optimal solution.

{\theorem{Let $\rho$ be continuous in $b$ and bounded by $\rho_{max}$, and $\epsilon>0$, then the root action values computed by $\rho$-POMCP$(\beta)$ converge asymptotically to $\epsilon$-optimal action values.}}

\begin{proof}
Since $\rho$ is bounded and the criterion is $\gamma$-discounted, the $\epsilon$-convergence allows reasoning with a finite horizon only, even though the problem horizon is infinite (with an infinite belief space).
  To do so, let us consider the tree of depth
  $\delta=\lceil
  \frac{\ln{\frac{\epsilon(1-\gamma)}{\rho_{max}}}}{\ln{\gamma}}
  \rceil$, and assume that the root belief estimate is exact.
  Due to UCB1, all nodes in that tree are visited infinitely often.
  Each belief-node $h$ then collects infinitely many particles
  ($\vert \beta \vert + 1$ at each visit) and, due to the use of particle
  filters applied from the root node, the probability distribution
  induced by cumulative bags $B(h)$ converges to the true
    belief state $b(h)$.
  Let us prove by induction that, at any depth $d\leq \delta$, the action
  value estimates are bounded by $\gamma^{-d} \epsilon$.
  Trivially, any node's value estimate is absolutely bounded by
  $\frac{\rho_{max}}{1-\gamma}$, so that,
  in particular, the bound is $\gamma^{-\delta} \epsilon$ at depth
  $\delta$ (\cf def. of $\delta$).
  Let us now assume that the induction property holds at depth
  $d \in \{1,\dots,\delta\}$.
  Then for any node $h$ at depth $d-1$,
  (i) due to $\rho$'s continuity, $\rho(b(h))$ is correctly estimated, \footnote{It is only required that the bias of the $\rho(B(h))$ estimator vanishes when the number of particles grows, which is guaranteed by $\rho$'s continuity.}
  (ii) since each action is infinitely selected, by induction, the action values converge to $\gamma^{d-1}\epsilon$-optimal values, and
  (iii) UCB1 selects the optimal action infinitely more often than other actions,
  thus $V(h)$ converges to a $\gamma^{d-1}\epsilon$-optimal
  value.
  \qed
\end{proof}

\subsection{$\rho$-POMCP$(\beta)$ variants}
\label{app:rhoPOMCP-variants}

\paragraph{Studied variants}

The current difference between POMCP and $\rho$-POMCP$(\beta)$ lies in the way particles are collected in order to estimate the reward obtained at each transition.
But value estimates are updated in the same way.
However, these updates can be improved, leading to several $\rho$-POMCP$(\beta)$ variants, since, during its execution, $\rho$-POMCP builds increasingly better estimates of the true belief states in visited belief nodes.
To do so, we first discuss what is computed by updates performed by POMCP (and vanilla $\rho$-POMCP$(\beta)$), and then present two variants we developed, which Bargiacchi had already proposed \cite{Bargiacchi16}.

\paragraph{Computations performed by POMCP}
If we ignore the node initialization in POMCP, then, when considering a node-action pair $ha$, the value stored in
$V(ha)$ averages, over $N(ha)$ samples/visits:
\begin{itemize}
\item $ \sum_{s\in\bag_{ha}} r(s,a)$:   the total reward over the states $s$ that were sampled while action
  $a$ was picked in $h$
  ($\bag_{ha}$ denotes this set of states);
\item
  $ \sum_{z\in \cZ_{ha}} \mathbf{1}_{N(h,a,z)\geq 1} Rollout(haz) $:   the total return of the rollouts generated from each observation $z$
  sampled after picking $a$ in $h$ (set $\cZ_{ha}$), where $N(h,a,z)$
  is the number of times action $a$ was followed by observation $z$ in
  node $h$ (while $N(haz)$ is the number of updates of node $haz$);
\item $\sum_{z\in \cZ_{ha}} \sum_{a'\in \cA} N(haza') V(haza')$:   the sum of the values of the children nodes, weighted by their visit
  counts (which also includes the rollouts performed from these nodes).
\end{itemize}
By introducing belief node value estimates $V(h)$, initialized with rollout values (for $N(h)=1$),
this leads to the following formulas:
\begin{align}
  V(h)
  & \gets \frac{1}{N(h)} \left[     Rollout(h) +
    \sum_{a' \in \cA} N(ha') V(ha')
    \right];
  \\
  V(ha)
  & \gets \frac{1}{N(ha)} \left[     \sum_{s\in\bag_{ha}} r(s,a) +
    \gamma \sum_{z\in \cZ_{ha}} N(haz)  V(haz)
    \right].
\end{align}

\paragraph{Last-reward-update $\rho$-POMCP$(\beta)$}
\label{sec:last-reward}
Thus, in POMCP, $V(ha)$ is a moving average that {\emph ``contains''} an estimate of
$r(b(h),a)$, that estimate being computed as
$\frac{\sum_{s\in\bag_{ha}} r(s,a)}{N(ha)}$.

In vanilla $\rho$-POMCP$(\beta)$, the sampled $r$ at a
current time step is replaced with an estimate of $\rho(b(h))$ as
$\rho(\hat{B}_{N(h)}(h))$, where $\hat{B}_{N(h)}(h)$ is
the cumulative bag after the first $N(h)$ visits.
In this case, $V(ha)$ includes thus (among other elements) an average of successive estimates (\ie, it computes
$\sum_{i=1}^{N(ha)} \rho(\hat{B}_{\phi(i)}(h),a)$, where $\phi(i)$ is the
$i^{th}$ visit of $h$ where $a$ was selected).
However, it would seem more appropriate to instead use the reward associated to the last estimated belief
$\rho(\hat{B}_{\phi(N(ha))}(h),a)$, which is a less biased estimate. 

This proposed variant, called \emph{last-reward-update $\rho$-POMCP$(\beta)$} (or {\em ``lru-$\rho$-POMCP$(\beta)$''}), fixes this rather easily by replacing the update of $V(ha)$ in Algorithm~\ref{alg:rhoPOMCP} by
\begin{align}
V(ha)
       & \gets
         \frac{N(ha)-1}{N(ha)} \left[ V(ha) - \rho^{prev}(h,a)\right] \\
      & \quad         + \rho(B(h),a) + \frac{1}{N(ha)} \gamma . \text{\simulateFct}(s' , haz, \depth + 1),
\end{align}
where
$\rho^{prev}(h,a)$ is the previous value of the reward when $ha$
was last experienced (thus needs to be stored).

\paragraph{Last-value-update $\rho$-POMCP$(\beta)$}

But then, $V(ha)$ also ``contains'' estimates of average rewards for
future time steps, which suffer from the same issue.
To fix this, \simulateFct should not return a sample return, but an
estimate of the average return.

The updates in the backpropagation step consist then in re-estimating all the values of the visited belief nodes by using the reward obtained at each transition and the initial rollout, which needs to have been previously  stored. This is done in \emph{last-value-update $\rho$-POMCP$(\beta)$} variant (or {\em ``lvu-$\rho$-POMCP$(\beta)$''}) through the following formulas:
\begin{align}
  V(h) & \gets  \frac{1}{N(h)} \left[ Rollout(h) + \sum_{a} N(ha) V(ha) \right], \\
  V(ha) & \gets \rho(B(h),a) + \frac{\gamma}{N(ha)} \sum_{z}  \left[  N(haz) .  V(haz) \right].
\end{align}

\section{EXPERIMENTS}
\label{sec:XPs}

\subsection{Benchmark Problems}

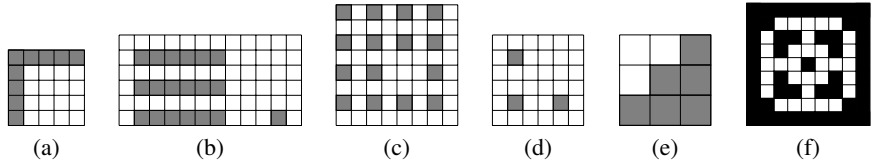
\begin{figure*}
  \center
  \begin{tabular}{cccccc}
    \scalebox{0.2}{        \begin{tikzpicture}
\draw [fill=gray!100] (0,0) rectangle (1,-5);
\draw [fill=gray!100] (0,0) rectangle (5,-1);
\draw (0,0) grid (5,-5);
\end{tikzpicture}
       }
    &
    \scalebox{0.2}{      \begin{tikzpicture}
\draw [fill=gray!100] (1,-1) rectangle (7,-2);
\draw [fill=gray!100] (1,-3) rectangle (7,-4);
\draw [fill=gray!100] (1,-5) rectangle (7,-6);
\draw [fill=gray!100] (10,-5) rectangle (11,-6);
\draw (0,0) grid (12,-6);
\end{tikzpicture}
     }
    &
    \scalebox{0.2}{      
\begin{tikzpicture}
\draw [fill=gray!100] (0,-0) rectangle (1,-1);
\draw [fill=gray!100] (0,-2) rectangle (1,-3);
\draw [fill=gray!100] (0,-4) rectangle (1,-5);
\draw [fill=gray!100] (0,-6) rectangle (1,-7);

\draw [fill=gray!100] (2,-0) rectangle (3,-1);
\draw [fill=gray!100] (2,-2) rectangle (3,-3);
\draw [fill=gray!100] (2,-4) rectangle (3,-5);
\draw [fill=gray!100] (2,-6) rectangle (3,-7);

\draw [fill=gray!100] (4,-0) rectangle (5,-1);
\draw [fill=gray!100] (4,-2) rectangle (5,-3);
\draw [fill=gray!100] (4,-6) rectangle (5,-7);

\draw [fill=gray!100] (6,-0) rectangle (7,-1);
\draw [fill=gray!100] (6,-2) rectangle (7,-3);
\draw [fill=gray!100] (6,-4) rectangle (7,-5);
\draw [fill=gray!100] (6,-6) rectangle (7,-7);
\draw (0,0) grid (8,-8);
\end{tikzpicture}
     }
    &
    \scalebox{0.2}{      \begin{tikzpicture}
  \draw [fill=gray!100] (1,-1) rectangle (2,-2);
  \draw [fill=gray!100] (4,-4) rectangle (5,-5);
  \draw [fill=gray!100] (1,-4) rectangle (2,-5);
  \draw (0,0) grid (6,-6);
\end{tikzpicture}
     }
    &
    \scalebox{0.4}{      \begin{tikzpicture}
\draw [fill=gray!100] (2,0) rectangle (3,-1);
\draw [fill=gray!100] (1,-1) rectangle (3,-2);
\draw [fill=gray!100] (0,-2) rectangle (3,-3);
\draw (0,0) grid (3,-3);
\end{tikzpicture}
     }
    &
    \scalebox{0.18}{      
\begin{tikzpicture}
  \draw [fill=black] (0,-0) rectangle (1,-9);
  \draw [fill=black] (8,-0) rectangle (9,-9);
  \draw [fill=black] (0,-0) rectangle (9,-1);
  \draw [fill=black] (0,-8) rectangle (9,-9);

  \draw [fill=black] (1,-1) rectangle (2,-2);
  \draw [fill=black] (7,-1) rectangle (8,-2);

  \draw [fill=black] (2,-2) rectangle (4,-3);
  \draw [fill=black] (5,-2) rectangle (7,-3);

  \draw [fill=black] (2,-3) rectangle (3,-4);
  \draw [fill=black] (6,-3) rectangle (7,-4);

  \draw [fill=black] (4,-4) rectangle (5,-5);

  \draw [fill=black] (2,-5) rectangle (3,-6);
  \draw [fill=black] (6,-5) rectangle (7,-6);

  \draw [fill=black] (2,-6) rectangle (4,-7);
  \draw [fill=black] (5,-6) rectangle (7,-7);

  \draw [fill=black] (1,-7) rectangle (2,-8);
  \draw [fill=black] (7,-7) rectangle (8,-8);

  \draw (0,0) grid (9,-9);
\end{tikzpicture}
     }
    \\
    (a)&(b)&(c)&(d)&(e)&(f) \\
  \end{tabular}

  \caption{The various cell configurations used for active localization problems; from left to right:
    (a) {\em MazeCross}, (b) {\em MazeLines}, (c) {\em MazeHole} and (d) {\em MazeDots};
    (e) corresponds to the cell configuration used for {\em GridX} and {\em GridNotX},
    and (f) the obstacle configuration for {\em SeekAndSeek}.}
  \label{fig:mazes}
\end{figure*}

POMDPs being a subclass of $\rho$-POMDPs, first benchmark problems we consider are the classical {\em Tiger}, {\em Tiger-Grid} and {\em Hallway2} problems \cite{littman95} (as per Cassandra's POMDP page) and instances of {\em Rock Sampling} \citep{SmiSim-uai04} with several grids of $n\times n$ cells, and $n$ rocks to sample (where $n$ equals 4, 6 and 8). A reward of $+100$ (resp. $-100$) is given for sampling a good (resp. bad) rock.

Then, a known issue with information-gathering problems is that a simple myopic strategy may often give very good results \citep{BonPeySab-ecai12,Satsangi17}.
In order to assess the proposed algorithms, we had to provide problems where myopic strategies encounter difficulties.

We first proposed the {\em Museum problem}, inspired by \citep{SatWhiSpa-tr15}, where an agent has to continuously localize a visitor in a toric grid environment ($4\times 4$ in our experiments).
The state corresponds to the visitor's unknown location.
At each time step, the visitor stays still with probability $0.6$, and moves to $1$ of his $4$ neighboring cells with probability $0.4$ (chosen uniformly).
The agent acts by activating a camera in any location.
It then receives a deterministic observation: \emph{``present''} if the visitor is at this location,
\emph{``close''} if s/he is in a neighboring cell, and
\emph{``absent''} if s/he is further away.
Additionally: the immediate reward corresponds to the negentropy of the belief: $\rho(b,a,b') = -H(b') = \sum_{s} b'(s). log(b'(s))$;
the initial belief $b_0$ is uniform over all cells;
and $\gamma=0.95$.
The interest of this problem lies in the large number of actions (one per cell).
We also used a variant, {\em Museum Threshold}, as proposed in Sec.~\ref{subsec:rhoPOMDP}, where the (non-continuous) reward is based on a threshold function on the belief state (with $\alpha = 0.8$) which is null in most of the belief space.

In {\em active-localization} problems,
an agent is in a toric grid with white and black cells.
At each step, it can move to a neighboring cell or observe the color of its current cell.
Active localization is difficult for myopic strategies as they see no (immediate) benefit in moving.
Additionally:
observations and transitions are deterministic;
$b_0$ is uniform over all cells;
the reward corresponds to the entropy difference between $b$ and $b'$:
$\rho(b,a,b')= - H(b')+H(b)$;
and $\gamma = 0.95$.
Several configurations have been studied (\cf Fig.~\ref{fig:mazes}):
(i) {\em MazeCross}, where black cells  make it easy to localize oneself;
(ii) {\em MazeLines},
where the agent cannot localize itself within the striped region and has to plan several steps ahead to look for the spot in the empty region;
(iii) {\em MazeHole},
which requires the agent to reason one step in advance to search for the missing black cell in this regular configuration;
and (iv) {\em MazeDots}, identical to MazeHole except that black cells are separated by several white cells.
{\em MazeDots\_nxn} corresponds to variations of the \textit{MazeDots} configuration where $n$ denotes the size of the grid.

{\em GridX} and {\em GridNotX} \citep{FehBufThoDib-nips18} differ from the first localization problems in that:
(i) moves (n,s,e,w) succeed with probability $0.8$, otherwise the agent
stays still;
(ii) denoting  $b_x$ (resp. $b_y$) the belief over the $x$ (resp. $y$)
coordinate, the reward function is  $\rho(b) = +\norm{b_x-\frac{1}{3} \mathbf{1}}_1$
(resp. $-\norm{b_x-\frac{1}{3} \mathbf{1}}_1$) for {\em GridX} (resp. {\em GridNotX}).
{\em GridNotX} is an interesting problem since the objective consists in maximizing uncertainty,
which can not be modeled with state-depend rewards and POMDP-IR.

In the {\em SeekAndSeek} problem, an object is lost in a (toric grid) maze with obstacles limiting the possible moves of the agent.
The agent can move in the maze and receives specific observations whenever the object is
on its location ({\em present}),
next to it ({\em close}), or
otherwise ({\em absent}).
The immediate reward received is the entropy difference over the possible object locations $\rho(b,a,b')=-H(b')+H(b)$.
The object being at a fixed location, a myopic agent surrounded by already visited cells has no incentive to explore any of them since they do not give any information.

We also addressed \citeauthor{ArayaThesis2013}'s {\em CameraClean} Diagnosis problem \cite{ArayaThesis2013}.
Here, a robot camera can be oriented to shoot one of four different zones and must find an object put in a fixed location.
Its state is its current target zone, and a boolean specifying whether its lens is clean or dirty.
As actions, the robot camera can reorient itself (deterministically) to the next zone, shoot the current zone, or clean its lens.
When the camera takes a picture, the probability to properly detect
the presence or absence of the object in the target zone is $0.8$ if
the lens is clean, and $0.55$ if dirty.
The reward is the entropy difference between starting and arrival
belief states $\rho(b,a,b') = -H(b')+H(b)$.

Finally, we proposed the \textit{LostOrFound} problem to consider non-convex rewards.
An agent is in a toric corridor with one colored cell.
It can move left or right (with probability 0.3 to stay), stay or change its status (\textit{lost} or \textit{found}). Then, it observes the color of its cell.
The reward depends on the agent's status.
If its status is \textit{found}, the agent tries to localize itself and $\rho(b,a,b') = H_{max}-H(b')$, where $H_{max} = log(\vert cells \vert)$ denotes the maximum entropy to ensure positive rewards.
If its status is \textit{lost}, it tries to get lost and receives $\rho(b,a,b') = 3.H(b')$.
To act optimally, the agent needs to change its status to \textit{lost} and to perform moving actions to lose information regarding its location.

All experiments were conducted on Intel Xeon Gold 6130 cores @ 2.1\,GHz with 512\,MB. Source code is available at {\scriptsize \url{https://gitlab.inria.fr/vthomas/ecai_2020_source}}.

\subsection{Results}

\paragraph{Influence of rollouts}

Preliminary experiments have been conducted with random rollouts.
For the same number of descents, $\rho$-POMCP$(\beta)$ with random rollouts (stopped when $\gamma ^{depth} < 0.01$) lasts between $5$ and $10$ times longer than without any rollout (setting the value of new nodes to $0$).
Moreover, the gain from using a pure random rollout policy was also rarely significant.
In some problems ({\em Tiger}), it even reduced the observed performance. We thus focus on $\rho$-POMCP$(\beta)$ and $\rho$-beliefUCT without rollouts.
Note that, due to implementation details,
  \citeauthor{Bargiacchi16}'s $\rho$-POMCP \cite{Bargiacchi16} may
  exhibit a different behavior than $\rho$-POMCP$(0)$.

\begin{table*}[ht]
  \captionsetup{justification=centering}
  {\caption{
  Results obtained when executing $200$ episodes of $40$ actions.
  $V$ is the averaged $\gamma$-discounted performance; $Err$ is the standard error; and $t(\text{s})$ is the average duration of one episode.
  $\rho$-MCTS algorithms were launched with a fixed number of 10\,000 descents before performing each action.
  Significantly best values are highlighted in bold.
  }\label{tab:basicResults}}
  \centerline{
    \resizebox{0.81\textwidth}{!}{
      \begin{tabular}{c X@{}Y@{}R@{} X@{}Y@{}R@{} X@{}Y@{}R@{} X@{}Y@{}R@{} X@{}Y@{}R@{}}
\toprule 
  & \multicolumn{3}{ c}{Random} &\multicolumn{3}{ c}{Look-ahead-1} &\multicolumn{3}{ c}{Look-ahead-3} &\multicolumn{3}{ c}{$\rho$beliefUCT} &\multicolumn{3}{ c}{$\rho$POMCP $\vert \beta \vert = 50$} \\ 
\cmidrule(lr){2-4} \cmidrule(lr){5-7} \cmidrule(lr){8-10} \cmidrule(lr){11-13} \cmidrule(lr){14-16}  
  Problem (UCB cst) &   $V$ & $Err$ & $t(s)$ &  $V$ & $Err$ & $t(s)$ &  $V$ & $Err$ & $t(s)$ &  $V$ & $Err$ & $t(s)$ &  $V$ & $Err$ & $t(s)$  \\ 
\cmidrule(lr){0-0}\cmidrule(lr){2-4} \cmidrule(lr){5-7} \cmidrule(lr){8-10} \cmidrule(lr){11-13} \cmidrule(lr){14-16}  
Tiger ($360$) & {-122.96} & {2.59} & {0.00} &  {\bf 1.80} & {0.13} & {0.02} &  {\bf 1.77} & {0.12} & {0.08} &  {\bf 1.92} & {0.13} & {6.44} &  {\bf 2.04} & {0.11} & {63.63} \\ 
\rowcolor[gray]{.9}[.5\tabcolsep]
CameraClean ($14$) & {0.39} & {0.01} & {0.00} &  {0.19} & {0.01} & {0.02} &  {\bf 0.82} & {0.02} & {0.07} &  {\bf 0.81} & {0.02} & {8.46} &  {\bf 0.81} & {0.02} & {64.59} \\ 
SeekAndSeek ($69.3$) & {-42.04} & {1.43} & {0.01} &  {-37.70} & {1.46} & {0.06} &  {-30.75} & {1.36} & {0.43} &  {\bf -23.25} & {1.16} & {32.73} &  {\bf -25.68} & {1.20} & {72.61} \\ 
\midrule 
\rowcolor[gray]{.9}[.5\tabcolsep]
Hallway2 ($1$) & {0.02} & {0.00} & {0.01} &  {0.03} & {0.00} & {0.62} &  {0.15} & {0.01} & {723.19} &  {\bf 0.21} & {0.01} & {46.17} &  {\bf 0.19} & {0.01} & {36.40} \\ 
TigerGrid ($1$) & {-1.73} & {0.15} & {0.01} &  {-0.08} & {0.04} & {0.22} &  {0.52} & {0.06} & {217.54} &  {\bf 1.93} & {0.05} & {20.48} &  {\bf 1.81} & {0.06} & {25.29} \\ 
\midrule 
\rowcolor[gray]{.9}[.5\tabcolsep]
Museum entropy ($1$) & {-26.31} & {0.23} & {0.02} &  {\bf -16.85} & {0.30} & {0.16} &  {-16.95} & {0.27} & {70.02} &  {\bf -16.09} & {0.30} & {24.11} &  {\bf -16.77} & {0.28} & {51.83} \\ 
Museum threshold ($1$) & {1.71} & {0.07} & {0.02} &  {6.30} & {0.16} & {0.18} &  {\bf 6.78} & {0.17} & {71.53} &  {\bf 6.58} & {0.17} & {28.63} &  {\bf 6.60} & {0.17} & {83.29} \\ 
\midrule 
\rowcolor[gray]{.9}[.5\tabcolsep]
ActivLocCross ($3.2$) & {1.38} & {0.04} & {0.01} &  {2.33} & {0.02} & {0.04} &  {\bf 2.58} & {0.01} & {0.32} &  {\bf 2.57} & {0.01} & {25.70} &  {\bf 2.57} & {0.01} & {39.99} \\ 
ActivLocLines ($4.3$) & {1.11} & {0.04} & {0.01} &  {1.99} & {0.04} & {0.08} &  {2.78} & {0.03} & {0.87} &  {\bf 2.99} & {0.03} & {72.46} &  {\bf 2.96} & {0.03} & {50.59} \\ 
\rowcolor[gray]{.9}[.5\tabcolsep]
ActivLocHole ($4.2$) & {0.92} & {0.03} & {0.01} &  {1.59} & {0.04} & {0.07} &  {\bf 2.36} & {0.04} & {0.79} &  {\bf 2.28} & {0.04} & {64.34} &  {\bf 2.29} & {0.04} & {60.18} \\ 
ActivLocDots ($3.6$) & {0.76} & {0.05} & {0.01} &  {1.56} & {0.05} & {0.05} &  {\bf 2.09} & {0.04} & {0.48} &  {\bf 2.15} & {0.04} & {41.61} &  {\bf 2.15} & {0.04} & {54.69} \\ 
\rowcolor[gray]{.9}[.5\tabcolsep]
ActivLocDots 8x8 ($4.2$) & {0.48} & {0.03} & {0.01} &  {1.15} & {0.04} & {0.06} &  {1.65} & {0.03} & {0.81} &  {\bf 1.77} & {0.04} & {67.12} &  {\bf 1.74} & {0.04} & {83.33} \\ 
ActivLocDots 10x10 ($4.6$) & {0.33} & {0.03} & {0.02} &  {0.73} & {0.04} & {0.10} &  {1.16} & {0.04} & {1.28} &  {\bf 1.34} & {0.05} & {93.83} &  {\bf 1.25} & {0.04} & {121.00} \\ 
\rowcolor[gray]{.9}[.5\tabcolsep]
ActivLocDots 12x12 ($5$) & {0.27} & {0.03} & {0.02} &  {0.60} & {0.04} & {0.12} &  {\bf 0.98} & {0.04} & {1.69} &  {\bf 0.98} & {0.05} & {130.12} &  {\bf 0.96} & {0.04} & {160.93} \\ 
\midrule 
GridX ($26$) & {16.69} & {0.14} & {0.01} &  {\bf 21.62} & {0.04} & {0.03} &  {\bf 21.72} & {0.04} & {0.34} &  {\bf 21.68} & {0.04} & {82.12} &  {\bf 21.73} & {0.04} & {448.01} \\ 
\rowcolor[gray]{.9}[.5\tabcolsep]
GridNotX ($26$) & {-16.91} & {0.15} & {0.01} &  {-4.57} & {0.20} & {0.04} &  {-3.95} & {0.15} & {0.37} &  {\bf -3.22} & {0.14} & {13.89} &  {\bf -3.04} & {0.14} & {70.26} \\ 
\midrule 
RockSampling44 ($100$) & {-20.95} & {4.15} & {0.01} &  {18.10} & {1.82} & {0.05} &  {99.12} & {4.49} & {2.09} &  {\bf 115.67} & {4.92} & {12.48} &  {\bf 109.15} & {4.76} & {61.34} \\ 
\rowcolor[gray]{.9}[.5\tabcolsep]
RockSampling66 ($100$) & {-5.66} & {1.70} & {0.02} &  {5.26} & {0.73} & {0.17} &  {\bf 96.04} & {4.32} & {10.08} &  {82.59} & {4.14} & {39.46} &  {78.21} & {3.74} & {107.09} \\ 
RockSampling88 v1($100$) & {-3.83} & {1.72} & {0.09} &  {6.33} & {1.06} & {1.11} &  {158.49} & {4.87} & {46.01} &  {\bf 206.05} & {5.46} & {107.79} &  {177.31} & {5.95} & {131.09} \\ 
\rowcolor[gray]{.9}[.5\tabcolsep]
RockSampling88 v2 ($100$) & {-4.28} & {1.20} & {0.05} &  {6.53} & {0.83} & {0.65} &  {\bf 140.38} & {4.88} & {49.04} &  {114.66} & {6.12} & {169.70} &  {104.53} & {5.99} & {178.04} \\ 
RockSampling88 v3 ($100$) & {-2.73} & {1.02} & {0.05} &  {4.79} & {0.98} & {0.63} &  {\bf 117.47} & {4.93} & {53.26} &  {86.21} & {3.83} & {170.42} &  {91.60} & {4.24} & {177.57} \\ 
\midrule 
\rowcolor[gray]{.9}[.5\tabcolsep]
LostOrFound ($35$) & {24.40} & {0.69} & {0.01} &  {31.23} & {0.00} & {0.02} &  {61.75} & {0.28} & {0.14} &  {\bf 62.97} & {0.21} & {53.84} &  {61.82} & {0.38} & {495.08} \\ 
LostOrFound ($100$) & {25.18} & {0.66} & {0.01} &  {31.23} & {0.00} & {0.03} &  {\bf 61.85} & {0.27} & {0.14} &  {31.38} & {0.15} & {7.08} &  {31.23} & {0.00} & {49.01} \\ 
\bottomrule 
\end{tabular}
     }
  }
\end{table*}

\paragraph{Comparison with myopic strategies}
Table~\ref{tab:basicResults} presents results comparing the purely \emph{Random} strategy, \emph{Look-ahead} strategies, $\rho$-beliefUCT, and $\rho$-POMCP$(\beta)$ on the benchmark problems.
The \emph {look-ahead-$H$} algorithms perform dynamic programming over all possible futures for a fixed finite horizon $H$ by using the complete $\rho$-POMDP model.
The pure myopic strategy, where the agent maximizes its immediate reward, corresponds to \emph{Look-ahead-1}, whereas \emph{Look-ahead-3} corresponds to anticipating all consequences 3 time-steps in advance\footnote{To prevent side-effects, when several actions share the same highest value, the performed action is randomly selected among them.}.
Both $\rho$-MCTS algorithms ($\rho$-POMCP$(\beta)$ and $\rho$-beliefUCT) use a fixed number of descents $nb_{descents}=10\,000$, without any rollout (setting the value of new nodes to $0$) and use a specific constant UCB for each problem  as specified in Tab.~\ref{tab:basicResults} (usually $(R_{max}-R_{min}) / (1-\gamma)$).
For $\rho$-POMCP$(\beta)$, \betasize $=50$ and importance sampling was used.

(a) Tab.~\ref{tab:basicResults} shows that \emph{Look-ahead-1} is close to the best value only on \emph{Tiger}, \emph{GridX} and \emph{Museum entropy}.
Regarding \emph{museum} problems, the myopic strategy is less efficient than \emph{look-ahead-3} in \emph{Museum Threshold} due to the sparsity of non-zero rewards.

(b) It is known that \emph{Look-ahead-1} often gives good results and usually constitute a very good baseline \cite{BonPeySab-ecai12,Satsangi17}.
But, we have also compared our approaches to the more challenging \emph{Look-ahead-3} strategy.
In this case, $\rho$-POMCP$(\beta)$ and $\rho$-beliefUCT give better results than \emph{Look-ahead-1} and similar results to \emph{Look-ahead-3}.
In most cases,
the difference between \emph{Look-ahead-1} and \emph{$\rho$-POMCP$(\beta)$} is significant.
\emph{$\rho$-POMCP$(\beta)$} also improves on \emph{Look-ahead-3} in
\emph{SeekAndSeek}, \emph{RockSampling44}, \emph{GridNotX}
\emph{ActiveLocLines}, \emph{Hallway2} and \emph{TigerGrid}.

(c) $\rho$-POMCP$(\beta)$ and $\rho$-beliefUCT provide the same results for this number of descents and take usually a lot more time than the \emph{Look-ahead} algorithms.
$\rho$-beliefUCT is usually faster than $\rho$-POMCP$(\beta)$, but this depends on the problem since computational costs of these two algorithms come from different operations.
In $\rho$-beliefUCT, this cost is due to the computation of exact beliefs when a new belief node is added.
In $\rho$-POMCP$(\beta)$, it is due to the generation of small bags $\beta$
at each transition.
That is why, whereas the time needed by $\rho$-POMCP$(\beta)$ is more regular (except for the yet-unexplained case of \emph{GridX}), the time needed by $\rho$-beliefUCT largely depends on $|\cS|$.
For instance, belief computation is quick in \emph{Tiger} or \emph{CameraClean}, but requires more time in \emph{Active Localization} problems, where belief states include many states with a non-zero probability. 
Sec.~\ref{sec:discussion} proposes tentative solutions to $\rho$-POMCP$(\beta)$'s high computational cost.

(d) In problems with larger state or observation spaces, like \emph{Hallway2} and \emph{TigerGrid}, $\rho$-MCTS algorithms are faster than \emph{Look-ahead-3} (by a factor between 8 and 20 depending on the problem) while achieving a higher performance.
It shows that $\rho$-MCTS algorithms manage to deal with problems even with a high branching factor (where \emph{Look-ahead-3} cannot compete) by focusing their descents on interesting branches.

(e) Finally, results from {\em Rocksampling} are difficult to analyze since they highly depend on the rock locations (different for $v1$, $v2$ and $v3$). The random action selection in {\em Look-head} when no reward is visible gives a good exploration policy for reaching rocks, whereas $\rho$-MCTS require more time to reach the interesting rewards (\cf Tab.~\ref{tab:timeBudget_1s_notapp}).

\begin{table*}[ht]
  \captionsetup{justification=centering}
  \caption{
  Influence of \betasize for a fixed time budget with a budget of $1$s per action (except last column with a budget of $10$s for reference).
  Each column corresponds to the results obtained for $200$ runs of $40$ actions by $\rho$-POMCP$(\beta)$ and several values of \betasize.
  $V$ is the average $\gamma$-discounted cumulative value, $Err$ the standard Error
  and $nb_{d}$ the number of descents performed.
  Significantly best values are highlighted in bold.
  }
  \label{tab:timeBudget_1s_notapp}
  \centerline{
    \resizebox{0.90\textwidth}{!}{
      \begin{tabular}{c X@{}Y@{}R@{} X@{}Y@{}R@{} X@{}Y@{}R@{} X@{}Y@{}R@{} X@{}Y@{}R@{} X@{}Y@{}R@{}}
\toprule 
  & \multicolumn{3}{ c}{$\vert \beta \vert = 0$} &\multicolumn{3}{ c}{$\vert \beta \vert = 1$} &\multicolumn{3}{ c}{$\vert \beta \vert = 5$} &\multicolumn{3}{ c}{$\vert \beta \vert = 10$} &\multicolumn{3}{ c}{$\vert \beta \vert = 100$} &\multicolumn{3}{ c}{$\vert \beta \vert = 10$, $t=10s$} \\ 
\cmidrule(lr){2-4} \cmidrule(lr){5-7} \cmidrule(lr){8-10} \cmidrule(lr){11-13} \cmidrule(lr){14-16} \cmidrule(lr){17-19}  
  Problem (UCB cst) &   $V$ & $Err$ & $nb_d$ &  $V$ & $Err$ & $nb_d$ &  $V$ & $Err$ & $nb_d$ &  $V$ & $Err$ & $nb_d$ &  $V$ & $Err$ & $nb_d$ &  $V$ & $Err$ & $nb_d$  \\ 
\cmidrule(lr){0-0}\cmidrule(lr){2-4} \cmidrule(lr){5-7} \cmidrule(lr){8-10} \cmidrule(lr){11-13} \cmidrule(lr){14-16} \cmidrule(lr){17-19}  
Tiger ($360$) & {-13.43} & {0.00} & {30945} &  {-13.43} & {0.00} & {29763} &  {\bf 2.14} & {0.10} & {19639} &  {\bf 2.12} & {0.12} & {13900} &  {0.38} & {0.12} & {2766} &  {\bf 1.95} & {0.13} & {102553} \\ 
\rowcolor[gray]{.9}[.5\tabcolsep]
CameraClean ($14$) & {0.28} & {0.02} & {31976} &  {0.25} & {0.02} & {28418} &  {\bf 0.77} & {0.02} & {19092} &  {\bf 0.78} & {0.02} & {13942} &  {\bf 0.79} & {0.02} & {3289} &  {\bf 0.80} & {0.02} & {92671} \\ 
SeekAndSeek ($69.3$) & {\bf -28.39} & {1.25} & {17342} &  {\bf -27.46} & {1.21} & {14574} &  {\bf -28.04} & {1.23} & {10877} &  {\bf -27.82} & {1.19} & {8505} &  {\bf -27.07} & {1.18} & {1740} &  {\bf -26.37} & {1.13} & {65322} \\ 
\midrule 
\rowcolor[gray]{.9}[.5\tabcolsep]
Hallway2 ($1$) & {0.19} & {0.01} & {30226} &  {0.18} & {0.01} & {23153} &  {0.19} & {0.01} & {17522} &  {0.21} & {0.01} & {14052} &  {0.18} & {0.01} & {2561} &  {\bf 0.25} & {0.01} & {106838} \\ 
TigerGrid ($1$) & {\bf 1.73} & {0.06} & {58236} &  {\bf 1.81} & {0.06} & {48902} &  {\bf 1.84} & {0.06} & {38756} &  {\bf 1.79} & {0.06} & {31980} &  {\bf 1.82} & {0.06} & {9641} &  {\bf 1.80} & {0.06} & {184639} \\ 
\midrule 
\rowcolor[gray]{.9}[.5\tabcolsep]
Museum entropy ($1$) & {\bf -16.51} & {0.29} & {29418} &  {\bf -16.81} & {0.27} & {24674} &  {\bf -16.74} & {0.29} & {19007} &  {\bf -16.20} & {0.29} & {15034} &  {-17.01} & {0.29} & {3052} &  {\bf -16.64} & {0.28} & {65596} \\ 
Museum threshold ($1$) & {6.15} & {0.17} & {31232} &  {6.15} & {0.17} & {26468} &  {\bf 6.23} & {0.18} & {19334} &  {\bf 6.59} & {0.18} & {14958} &  {\bf 6.70} & {0.16} & {3196} &  {\bf 6.41} & {0.18} & {81593} \\ 
\midrule 
\rowcolor[gray]{.9}[.5\tabcolsep]
ActivLocCross ($3.2$) & {\bf 2.58} & {0.01} & {20602} &  {\bf 2.59} & {0.01} & {18893} &  {\bf 2.56} & {0.01} & {15391} &  {\bf 2.57} & {0.01} & {12475} &  {\bf 2.57} & {0.01} & {2837} &  {\bf 2.58} & {0.01} & {99308} \\ 
ActivLocLines ($4.3$) & {2.89} & {0.03} & {11859} &  {2.89} & {0.03} & {11188} &  {2.90} & {0.03} & {9208} &  {\bf 2.90} & {0.03} & {7697} &  {2.87} & {0.03} & {1693} &  {\bf 2.98} & {0.03} & {58659} \\ 
\rowcolor[gray]{.9}[.5\tabcolsep]
ActivLocHole ($4.2$) & {\bf 2.36} & {0.04} & {12737} &  {\bf 2.39} & {0.04} & {12025} &  {2.26} & {0.04} & {9556} &  {2.27} & {0.04} & {8152} &  {\bf 2.33} & {0.04} & {1709} &  {\bf 2.34} & {0.04} & {57199} \\ 
ActivLocDots ($3.6$) & {\bf 2.15} & {0.04} & {15897} &  {\bf 2.18} & {0.04} & {14965} &  {\bf 2.16} & {0.04} & {12031} &  {\bf 2.21} & {0.04} & {9288} &  {\bf 2.18} & {0.04} & {2035} &  {\bf 2.14} & {0.04} & {70543} \\ 
\midrule 
\rowcolor[gray]{.9}[.5\tabcolsep]
GridX ($26$) & {\bf 21.69} & {0.04} & {44487} &  {21.67} & {0.04} & {35916} &  {\bf 21.79} & {0.04} & {23823} &  {21.67} & {0.04} & {17583} &  {21.68} & {0.04} & {3718} &  {21.61} & {0.04} & {101912} \\ 
GridNotX ($26$) & {-3.46} & {0.15} & {40786} &  {\bf -3.30} & {0.14} & {33989} &  {-3.54} & {0.14} & {23351} &  {\bf -3.17} & {0.13} & {16605} &  {\bf -3.41} & {0.14} & {3627} &  {\bf -3.01} & {0.15} & {123545} \\ 
\midrule 
\rowcolor[gray]{.9}[.5\tabcolsep]
RockSampling44 ($100$) & {\bf 114.00} & {4.68} & {21967} &  {\bf 110.78} & {5.08} & {18991} &  {\bf 106.79} & {4.69} & {13059} &  {\bf 109.00} & {4.87} & {9952} &  {103.89} & {4.33} & {2198} &  {\bf 118.46} & {4.59} & {66347} \\ 
RockSampling66 ($100$) & {\bf 110.12} & {4.37} & {14033} &  {\bf 108.86} & {4.38} & {11981} &  {\bf 103.77} & {4.37} & {8915} &  {88.08} & {4.13} & {6416} &  {50.00} & {3.98} & {1112} &  {\bf 109.88} & {4.02} & {55186} \\ 
\rowcolor[gray]{.9}[.5\tabcolsep]
RockSampling88 v1($100$) & {172.69} & {6.10} & {6228} &  {155.01} & {6.11} & {5195} &  {152.48} & {5.57} & {3791} &  {123.84} & {5.32} & {2946} &  {50.02} & {4.42} & {542} &  {\bf 188.54} & {5.07} & {23907} \\ 
RockSampling88 v2($100$) & {130.21} & {5.85} & {4615} &  {113.39} & {5.67} & {4378} &  {108.54} & {5.28} & {2990} &  {99.43} & {5.26} & {2286} &  {41.93} & {3.89} & {420} &  {\bf 149.96} & {5.38} & {26922} \\ 
\rowcolor[gray]{.9}[.5\tabcolsep]
RockSampling88 v3($100$) & {116.15} & {5.03} & {5732} &  {118.31} & {5.12} & {4634} &  {109.51} & {4.59} & {3181} &  {102.24} & {4.67} & {2403} &  {68.04} & {4.04} & {466} &  {\bf 171.84} & {4.56} & {24677} \\ 
\midrule 
LostOrFound ($35$) & {31.23} & {0.00} & {32734} &  {31.23} & {0.00} & {27317} &  {31.87} & {0.35} & {19088} &  {39.68} & {1.12} & {14826} &  {\bf 59.50} & {0.32} & {4019} &  {37.92} & {0.96} & {74955} \\ 
\rowcolor[gray]{.9}[.5\tabcolsep]
LostOrFound ($100$) & {50.32} & {0.79} & {35012} &  {59.86} & {0.22} & {33141} &  {58.27} & {0.27} & {22519} &  {55.42} & {0.37} & {15824} &  {31.23} & {0.00} & {3462} &  {\bf 63.05} & {0.16} & {114450} \\ 
\bottomrule 
\end{tabular}
     }
  }
\end{table*}

\paragraph{Results with fixed time-budget}
Table~\ref{tab:timeBudget_1s_notapp} presents results regarding the influence of \betasize with a fixed time-budget of $1$\,s by action (except last column).
When \betasize increases, the number of descents performed by $\rho$-POMCP$(\beta)$ naturally decreases (since trajectory generation is slower),
but this has no clear impact on the $\gamma$-discounted cumulated value.
Several problems exhibit different behaviors.
In {\em CameraClean} and {\em Tiger}, there is a significant performance gap between \betasize$=0$ and \betasize$=5$.
This might come from the highly stochastic observation process which requires better belief state estimates to act correctly.
In \emph{Rocksampling}, small values of \betasize give better results.
This may be due to overestimation which favors exploitation, and more descents allowing to reach a greater depth.
Finally, results obtained with \textit{LostOrFound} need to be commented in detail. With a UCB constant of 35, $\rho$-POMCP$(\beta)$ needs a large bag to give good results.
In this problem, convex rewards (\textit{found} status) are compared with concave rewards (\textit{lost} status).
However, when we try to maximize information (convex reward), a poorly estimated belief lead to an optimistic return and $\rho$-POMCP$(\beta)$ is attracted by the corresponding branch.
On the contrary, with concave rewards, the estimated return is low and will be an incentive not to explore this branch anymore.
Both these effects lead $\rho$-POMCP$(\beta)$ to fail to find a good policy when \betasize is low.
When the UCB constant increases, this effect disappears due to favored exploration.
In (almost) all cases, when the time-budget increases (like 10\,s in the last column), $\rho$-POMCP$(\beta)$ manages to generate the highest cumulated return (higher than {\em Look-ahead-3} from Tab.~\ref{tab:basicResults}).

\paragraph{Comparison between rejection and importance sampling (IS)}
We have conducted experiments with $\rho$-POMCP$(\beta)$ algorithms to test the interest of using importance sampling instead of rejection sampling. All experiments were conducted with 200 runs of 40 actions, 10\,000 descents per action, $\vert \beta \vert=20$, and UCB constants from previous tables.
The cumulated values are the same and rejection sampling requires approximatively 20$\%$ more time on small problems (the table is not presented in this article).
However, for problems with larger observation spaces or with highly stochastic observation process, like {\em TigerGrid}, {\em Hallway2} or {\em Museum} problems, rejection sampling requires much more time (from to 2 to 8 times more than IS, \ie, from 48\,s to 381\,s for {\em TigerGrid} problem) because of a high reject rate.

\paragraph{Results with proposed variants}
To speed up convergence, {$lru$-$\rho$-POMCP$(\beta)$ and $lvu$-$\rho$-POMCP$(\beta)$ variants (\cf Sec.~\ref{app:rhoPOMCP-variants})} have been investigated replacing moving averages by up-to-date estimates, which are less biased in our setting.
However, up to now, experiments with fixed time budgets (100\,ms, 200\,ms, 1\,s and 10\,s) have not shown any significant improvement.

\section{DISCUSSION}
\label{sec:discussion}

In this article, we proposed two algorithms, $\rho$-POMCP$(\beta)$ and $\rho$-beliefUCT, to address $\rho$-POMDPs without constraints on the reward function.
$\rho$-POMCP$(\beta)$ (and trivially $\rho$-beliefUCT) is proved to asymptotically converge when $\rho$ is continuous and bounded.
$\rho$-MCTS algorithms are thus particularly useful when considering non-convex non-Lipschtiz-continuous rewards which cannot be addressed by previous approaches (like \citep{AraBufThoCha-nips10}, \citep{FehBufThoDib-nips18} or \citep{SatWhiSpa-tr15}).

Conducted experiments show that both algorithms give better results than the proposed baseline.
The difference between $\rho$-POMCP$(\beta)$ and original $\rho$-POMCP ($\beta=0$ in Table \ref{tab:timeBudget_1s_notapp}) is less significant.
However, we proposed problems (like \textit{LostOrFound}) where the use of a particle filter generates better results with a fixed-time budget (due to poorly estimated beliefs when using small particle bags).

This advantage of $\rho$-POMCP$(\beta)$ over original $\rho$-POMCP would probably increase with a better use of the available time budget.
We have observed that $\rho$-POMCP$(\beta)$ is time consuming due to the many sampled particles.
One promising direction would be to save time by letting a cumulative bag $B(h)$ grow sub-linearly (rather than linearly) with the number of visits $N(h)$, \ie, {\bf by considering dynamic \betasize}.
This will require modifying the way particles are generated, which is
no simple task since particles need to be propagated from the root to
the end of each trajectory.

Another issue that even concerns vanilla POMCP and particle filtering is how to deal with
{\bf unexpected transitions}.
If, after actually performing an action $a$, the actual observation $z$ has been rarely sampled (if at all), the new root may come with a poor belief estimate or not exist at all, so that the online computations will give poor results if they can be run at all.
In this case, performing particle filtering from the root belief's parent might provide a new root belief but with a high computational cost.
Another complementary improvement would be to regularly sample particles from the initial POMDP belief to add new particles in the current root and prevent particle depletion during episodes.

  Finally, as suggested by \citeauthor{SunKoc-corr18}
  \cite{SunKoc-corr18,SunKoc-icaps18}, one could build on their
  (related) work to address information-oriented control problems with
  {\bf continuous states/actions/observations}, \eg, in a robotic context.

\ack 
Experiments presented in this paper were carried out using the Grid'5000 testbed, supported by a scientific interest group hosted by Inria and including CNRS, RENATER and several Universities as well as other organizations (see \url{https://www.grid5000.fr}).

We would also like to thank Eugenio Bargiacchi, as well as the anonymous reviewers, for their feedback.

\bibliographystyle{ecainat-plain}

\ifdefined\bibliotex

\else

\fi

\end{document}